%% file: mypaper.tex
\documentclass[letterpaper]{article} % DO NOT CHANGE THIS
\usepackage{aaai25}  % DO NOT CHANGE THIS
\usepackage{times}  % DO NOT CHANGE THIS
\usepackage{helvet}  % DO NOT CHANGE THIS
\usepackage{courier}  % DO NOT CHANGE THIS
\usepackage[hyphens]{url}  % DO NOT CHANGE THIS
\usepackage{graphicx} % DO NOT CHANGE THIS
\usepackage{natbib}  % DO NOT CHANGE THIS AND DO NOT ADD ANY OPTIONS TO IT
\usepackage{caption} % DO NOT CHANGE THIS AND DO NOT ADD ANY OPTIONS TO IT
\usepackage{algorithm}
\usepackage{algorithmicx}
\usepackage{newfloat}
\usepackage{graphicx}
\usepackage{listings}
\usepackage{algpseudocode}
\usepackage{algorithm}
\usepackage{algpseudocode}
\usepackage{amssymb}

\DeclareCaptionStyle{ruled}{labelfont=normalfont,labelsep=colon,strut=off} % DO NOT CHANGE THIS
\floatstyle{ruled}
\newfloat{listing}{tb}{lst}{}
\floatname{listing}{Listing}
\title{Breaking Event Rumor Detection via Stance-Separated Multi-Agent Debate}
\nocopyright
\author {
    Mingqing Zhang\equalcontrib\textsuperscript{\rm 1,\rm 2},
    Haisong Gong\equalcontrib\textsuperscript{\rm 1,\rm 2},
    Qiang Liu\textsuperscript{\rm 1,\rm 2}\thanks{Corresponding Author},
    Shu Wu\textsuperscript{\rm 1,\rm 2},
    Liang Wang\textsuperscript{\rm 1,\rm 2}
}
\affiliations {
    \textsuperscript{\rm 1}New Laboratory of Pattern Recognition (NLPR)\\
State Key Laboratory of Multimodal Artificial Intelligence Systems\\
Institute of Automation, Chinese Academy of Sciences\\
    \textsuperscript{\rm 2}School of Artificial Intelligence, University of Chinese Academy of Sciences\\
    mingqing.zhang@cripac.ia.ac.cn,
    gonghaisong2021@ia.ac.cn, \{shu.wu, qiang.liu, wangliang\}@nlpr.ia.ac.cn
}

\usepackage{bbding}
\usepackage{color}
\usepackage{amsfonts,amssymb,amsmath} 
\usepackage{xspace}
\usepackage{booktabs}
\usepackage{multirow}
\usepackage[many]{tcolorbox}    	% for COLORED BOXES (tikz and xcolor included)
\newcommand{\themodel}{S2MAD\xspace}
\begin{document}
\maketitle

\begin{abstract}
The rapid spread of rumors on social media platforms during breaking events severely hinders the dissemination of
the truth. Previous studies reveal that the lack of annotated
resources hinders the direct detection of unforeseen breaking events not covered in yesterday’s news. Leveraging large
language models (LLMs) for rumor detection holds significant promise. However, it is challenging for LLMs to provide comprehensive responses to complex or controversial
issues due to limited diversity. In this work, we propose the
Stance Separated Multi-Agent Debate (\themodel) to address
this issue. Specifically, we firstly introduce Stance Separation, categorizing comments as either supporting or opposing the original claim. Subsequently, claims are classified as
subjective or objective, enabling agents to generate reasonable initial viewpoints with different prompt strategies for
each type of claim. Debaters then follow specific instructions
through multiple rounds of debate to reach a consensus. If a
consensus is not reached, a judge agent evaluates the opinions
and delivers a final verdict on the claim’s veracity. Extensive
experiments conducted on two real-world datasets demonstrate that our proposed model outperforms state-of-the-art
methods in terms of performance and effectively improves
the performance of LLMs in breaking event rumor detection.

\end{abstract}

\input{intro0}

\input{relatedwork}

\input{method}
\input{experiment}

\input{conclusion}
\bibliography{mypaper}
\end{document}

%% file: intro0.tex
\section{Introduction}

With the widespread use of social media platforms like Twitter and Weibo,  the occurrence of breaking events has provided fertile ground for the spread of rumors. During these events, information asymmetry, emotional intensity, and a lack of official updates often lead to controversy, further fueling rumor dissemination. These controversies heighten information confusion, enabling rumors to spread rapidly~\cite{knapp1944psychology,yang2022coarse}.

Traditional rumor detection methods depend on large corpora gathered from public sources to train models, which are then fine-tuned on target data to adapt to specific domains~\cite{bian2020rumor,qourbani2023toward,gong2024heterogeneous,cui2023kagn,lin2023zero,lin2024towards,zhang2024t3rd}.  However, these methods face significant challenges in Breaking Event Scenarios (BES), which are characterized by their suddenness, uniqueness, and controversy. This difficulty arises because breaking events occur abruptly, making it challenging to gather the necessary reliable datasets in real-time. This consequently limits their effectiveness in sudden, fast-paced events where up-to-date and accurate data are crucial.

Recent successes of Large Language Models (LLMs) in various reasoning tasks can be attributed to their extensive knowledge base and generalization capabilities~\cite{nam2024using}. Given these strengths, LLMs hold significant potential in the realm of rumor detection, especially under the challenging conditions typical of BES. In light of this, we propose leveraging the generative capacities and comprehensive human knowledge embedded within LLMs to address the challenges of rumor detection in these scenarios. Instead of relying solely on individual LLMs, which may lack diversity and robust validation mechanisms, and thus struggle with providing comprehensive and accurate responses to complex or controversial issues, we employ a multi-agent debate mechanism. This approach integrates various agents' viewpoints to offer more comprehensive and diverse analyses. By mimicking human debates, we prompt LLM agents to act as different debaters. These agents, with their extensive knowledge and reasoning skills, can thoroughly explore and utilize the limited resources available during BES.

To elaborate on the proposed solution, we introduce Stance Separated Multi-Agent Debate (\themodel), designed explicitly to counter the unique challenges of rumor detection on social media during BES. We first introduce Stance Separation, which meticulously categorizes comments on claims into groups that either support or oppose the original claim. The goal is to create a clear division of opinions, reflecting the varied perspectives typical in controversial debates. Based on whether the initial claim is subjective or objective, each debater is initialized with corresponding instructions to ensure they can clearly articulate and defend their assigned stance. Specific debate instructions are crafted for each debater to adhere to through several rounds of debating. The aim of these debates is to reach a consensus based on the differing viewpoints. If the debaters cannot arrive at a consensus through their rounds of contention, we design a judge agent to take both debaters' opinions into consideration and render a final verdict on the claim’s veracity. This comprehensive approach ensures that the limited information available in BES is maximally utilized, leveraging the LLMs' abilities to mirror human critical thinking and debate dynamics.

 Our main contributions can be summarized as follows:
\begin{itemize}
    \item We propose the Multi-Agent Debate for Rumor Detection (\themodel) approach, which integrates various agents' viewpoints to provide comprehensive analyses. This mechanism allows models to correct and validate each other, improving overall accuracy and reliability.
    \item The \themodel approach separates responses to claim into supportive or opposing categories. This helps agents take different stances during debates and thoroughly analyze the post's veracity.
    \item Extensive experiments on Twitter-COVID19 and Weibo-COVID19 
 datasets validate the effectiveness of \themodel, showing improved performance in social media rumor detection during breaking events.
\end{itemize}

%% file: relatedwork.tex
\section{Related Work}
\subsection{Rumor Detection}
Rumor detection on social media is a critical text mining task
aimed at identifying and mitigating the spread of misinformation. Researchers employ diverse methods to enhance rumor detection, crucial for maintaining information integrity
on social media~\cite{takahashi2012rumor,cai2014rumors,zhang2015automatic,sampson2016leveraging,pathak2020analysis}.
In recent years, deep learning has advanced rumor detection by automatically extracting features from text and propagation patterns. Recurrent Neural Networks (RNNs) and their variants have been widely used in this research~\cite{ma2015detect,ma2016detecting,ma2018rumor,naumzik2022detecting}.
Researchers have developed new methods to analyze the complex structure of rumor propagation, with Graph Neural Networks (GNNs) being particularly significant. GNNs model the structured information in rumor spread, offering deeper insights into its dynamics~\cite{wu2020rumor,xu2022evidence,zhang2023rumor,wu2023adversarial,tao2024semantic}. As Large Language Models (LLMs) have been widely applied to various tasks, some researchers are now exploring their application in detecting misinformation~\cite{li2024large,liu2024can,nan2024let,yan2024enhancing,gong2024navigating}.  However,
these methods have not yet fully explored the potential of LLMs in breaking events scenarios. Our research primarily focuses on this aspect, aiming to uncover the application
value of LLMs in these dynamic and challenging environments.

\subsection{Multi-agent Debate}
Recent research has explored how to utilize collaborative debates between agents, either from a single model or multiple different models, to enhance reasoning capabilities for specific tasks~\cite{chan2023chateval,du2023improving,liang2023encouraging,wang2024rethinking}. 
\citet{du2023improving} proposes a method where each agent provides its own answer to the same question and then refines it based on the responses from other agents.
\citet{liang2023encouraging, wang2024rethinking} initialized agents with different viewpoints and had them engage in in-depth debates, stimulating deeper analysis and
resulting in a more accurate and comprehensive reasoning process.
Additionally, \citet{zhang2024enhancing} employs Multi-LLMs Collaborated Reasoning, using different LLMs as agents to check prediction rationality from various aspects and aggregating their responses through debates, to enhance the capability of video-based human-object interaction detection. 
\citet{fang2024counterfactual} presets the stances of LLMs and compels them to generate justifications for predetermined answers. By engaging in counterfactual debates with a skeptical critic, this approach overrides the inherent biases of LLMs to achieve hallucination elimination. 
In this work, we introduced multiple agents and synthesizing their insights and judgments to deeply explore the application of multi-agent debate in social media rumor detection during breaking event scenarios.

%% file: method.tex
\newtcolorbox{mybox}{
    enhanced,
    breakable,
    boxrule = 1.0pt,
    rounded corners,
    colback = sub,
    left = 1mm,
    right = 1mm,
    top = 1mm,
    bottom = 1mm,
    before skip = 1.6mm,
    % after skip = -3mm,
    arc = 0pt   % corners roundness
}
\definecolor{sub}{HTML}{eeeeee}

\section{Method}
\begin{figure*}[t]
\setlength{\abovecaptionskip}{0pt}  % reduce gap
\setlength{\belowcaptionskip}{0pt}  % reduce gap
   \begin{center}
%   \vspace{-4mm}
\includegraphics[width=1\textwidth]{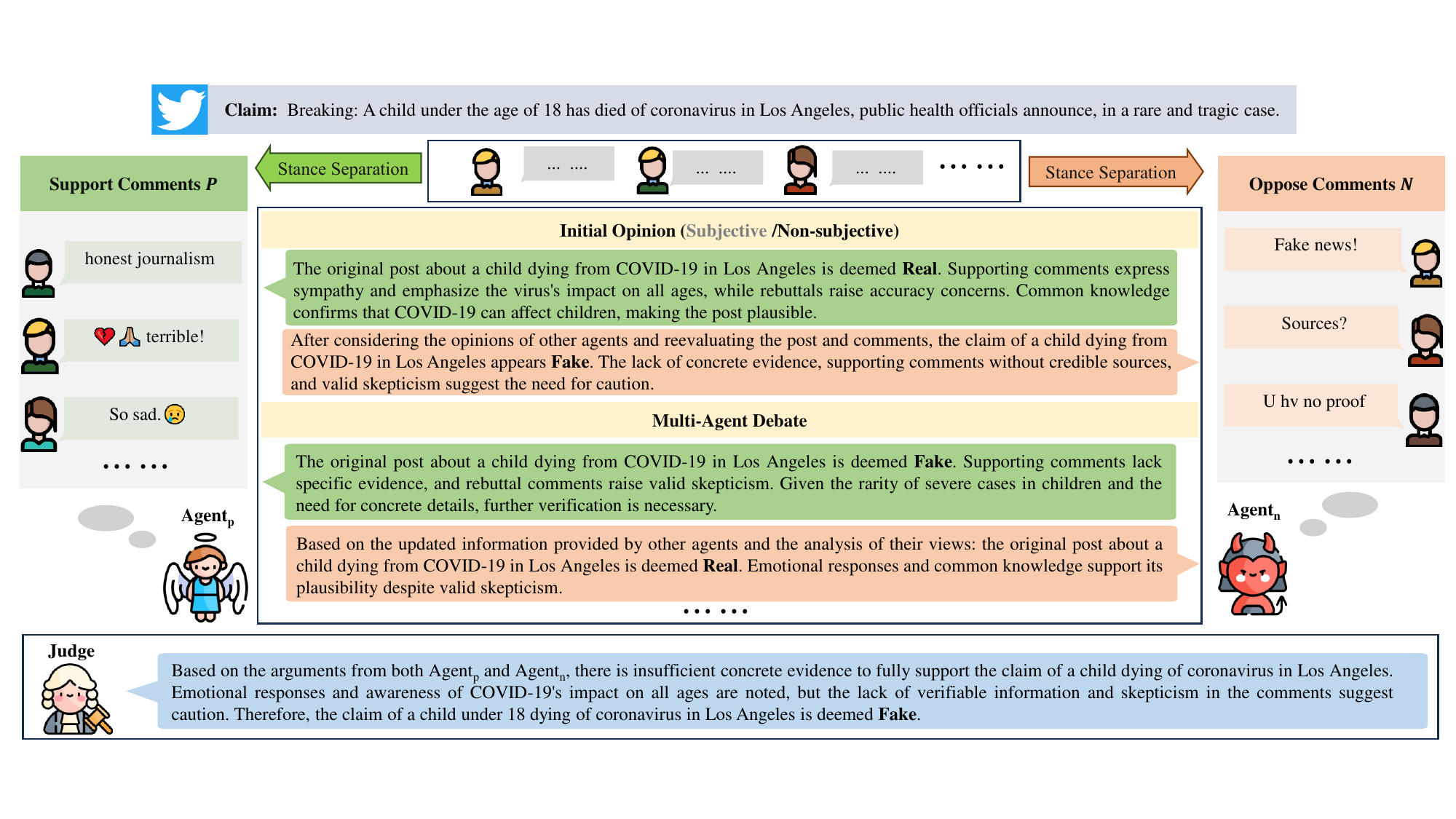}
   \end{center}
   \vspace{-0.7cm}
   % \caption{Architecture of \themodel. We first categorize the comments based on their stance towards the claim into supporting comments $P$ and opposing comments $N$. Furthermore, we classify the claim based on whether it is a subjective expression or not, and design corresponding guiding prompts for different types of claim. Utilizing LLMs as agents, we evaluate the truthfulness of the claim based on these different stances and generate an initial opinion. Subsequently, we conduct a multi-agent debate. If the agents do not reach a consensus after $M$ rounds, a judge is designed to derive the final prediction.} 

   \caption{Given an input claim and its related comments, we first categorize the comments based on
their stance towards the claim into supporting comments $P$ and opposing comments $N$. Furthermore, we classify the claim
based on whether it is a subjective expression or not, and design corresponding guiding prompts for different types of claim
(subjective and non-subjective). Utilizing LLMs as agents, we evaluate the truthfulness of the claim based on these different
stances and generate an initial opinion. Subsequently, we conduct a multi-agent debate. If the agents do not reach a consensus
after a maximum of $M$ rounds, a judge is designed to derive the final prediction.}
   \label{fig:model}
\end{figure*}

 Generally, \themodel consists of three key components: stance separation, initial opinion generation, and multi-agent debate. The overall framework is shown in Figure \ref{fig:model}. For a given social media claim and its comments, we perform stance separation based on the attitudes expressed in the comments, resulting in a support stance comment set P and an oppose stance comment set N. 
 % Further, we classify the claim based on whether it is a subjective expression
 % , and designed corresponding guiding prompts for different types of claims (subjective claims and non-subjective claims). 
 Subsequently, we treat LLM agents as debaters (referred to as $D_p$ and  $D_n$) and feed each debater with different stance comment set to initialize their default preference to the truthfulness of the claim with specific designed prompt according to their subjectivity. At last, we conduct a multi-agent debate process. If the debate process reaches the maximum number of rounds $M$ without consensus between the two debaters, we designed a judge to attend to each debater's opinion and conclude a final prediction.
 Algorithm \ref{alg} illustrates the detailed process of \themodel.
 %Using LLMs as agents, we assess the veracity of the claim based on comments from these differing stances to generate an initial opinion. 
 % We have a total of two agents, referred to as debaters, denoted as $D_p \text{and} D_n$.

\subsection{Problem Formulation} The task of rumor detection involves distinguishing between true and false events (claims), typically modeled as a binary text classification problem. In our scenario, given a claim $c$ along with its corresponding set of comments $\mathcal{T}(c)=\{(x_0,t_0),(x_1,t_1),\cdots,(x_n,t_n)\}$ , where $x_i$ represents the textual content of $i$th comment and $t_i$ indicates the relative posting time compared to $c$. The objective is to predict the label $y$ as either \texttt{True} or \texttt{False} based on the features of the claim itself and the set of comments.

\subsection{Stance Separation}
In a debate process, it is essential for each debater to have their own stance as a foundation for the following debate process. To initiate the stance for each debater based on the comments, we first need to provide each agent with a suitable subset of all comments $\mathcal{T}(c)$. Building upon previous research~\cite{wang2024explainable}, we aim to obtain two subsets from $\mathcal{T}(c)$ that representing contrasting stances. One subset comprises comments that support the claim, while the other consists of comments that oppose the claim. It is important to note that comments not aligning with common sense are excluded from this process.

To achieve this, we employ a LLM as a scorer, assigning positive scores in the range $(0.0,1.0]$ to comments supporting the claim and negative scores in the range $[-1.0,0.0)$ to those opposing the claim. Comments that do not adhere to common sense are assigned a score of $0.0$. This scoring process can be mathematically expressed as:
\begin{equation}\label{eq:score}
s_i = \text{LLM} (p_{ss},c,x_i), \quad s_i \in [-1,1] \end{equation}
where $s_i$ represents the score of the $i$th comment, $p_{ss}$ is the prompt template used for scoring:
\begin{mybox}
\textbf{Instruction}: First, check whether the comment conforms to common knowledge. If it does not conform to common sense, the score is $0.0$. Does the comment support that the source post is real? If it supports, it will be scored from the range $(0.0, 1.0]$ according to the degree of support. The higher the score, the stronger the support. If it does not support, it will be scored from the range $[-1.0, 0.0)$ according to the degree of opposition. The stronger the degree of opposition, the lower the score.\\
\textbf{Claim}: [Claim]\\
\textbf{Comment}: [Comment]\\
Now, output your answer strictly in the following format:\\
\{``Reason": ``", ``Score": ``"\}\\
Do not output any irrelevant content.

\end{mybox}

Typically, comments on a claim $c$ can be plentiful, and to ensure a clear stance, we selectively choose the top $k$ comments that are most supportive and most opposing to form the support stance comment set $P$ and oppose stance comment set $N$, mathematically expressed as:
\begin{equation}\label{eq:topk}
\begin{aligned}
    P &= \{x_i | s_i \in\text{topk}(\{s_i | s_i > 0\},k)\}\\
    N &= \{x_i | s_i \in \text{topk}(\{-s_i | s_i < 0\},k)\}
\end{aligned}
\end{equation}
where \text{topk} is a function that selects $k$ highest scores.

\iffalse
\subsection{Subjectivity Classification}
On social media, claims generally fall into two main categories: subjective statements and objective expressions. Objective sentences involve facts, data, and verifiable information, while subjective sentences express personal thoughts, such as opinions, evaluations, and emotions. These sentences can convey positive or negative sentiments. According to previous research~\cite{liu2024can}, proper guidance can enable agents to more effectively capture clues for reasoning. Subjective expressions and objective statements differ significantly in terms of language and information. By distinguishing between these two types, more targeted prompts can be designed for each type, thereby improving the accuracy of agents' judgments.
Specifically, we leverage the powerful reasoning capabilities of LLMs to classify whether a claim is a subjective expression from the user, and accordingly design subjectivity prompt template:
\begin{mybox}
\textbf{Instruction}: Does the claim only express the personal opinions of the Twitter user?\\
\textbf{claim}: [claim]\\
 Please answer `Yes' or `No'.
\end{mybox}
By categorizing the claim into subjective expressions and objective statements, we can fully utilize the reasoning and sentiment analysis capabilities of agents, thereby enhancing the accuracy and efficiency of rumor detection.
\fi

\subsection{Initial Opinion Generation}

We need to guide agents in analyzing claim content for rumor detection and use the support stance comment set $P$ and oppose stance comment set $N$ obtained above to generate initial opinions with different stances for each agent.
According to previous research~\cite{liu2024can}, proper guidance can enable agents to more effectively capture reasoning clues. On social media, claims generally fall into two main categories: subjective claims and objective claims. Objective claims involve facts, data, and verifiable information, while subjective claims express personal thoughts, such as opinions, evaluations, and emotions. These claims can convey positive or negative sentiments. Given the significant differences in language and information between subjective claims and objective claims, distinguishing between these types allows for more targeted prompts, thereby improving the accuracy of agents' judgments.

Specifically, we leverage the powerful reasoning capabilities of LLMs to classify whether a claim is a subjective expression from the user and accordingly design subjectivity prompt template:
\begin{mybox}
\textbf{Claim}: [Claim]\\
Does the claim only express the personal opinions of the user? Please answer Yes or No.
\end{mybox}
By categorizing the claim into subjective expressions and objective statements, we can fully utilize the reasoning and sentiment analysis capabilities of agents, thereby enhancing the accuracy and efficiency of rumor detection. Next, we designed prompts to let agents generate their own initial opinions respectively.
Specifically, we designed corresponding guiding prompts for different types of claims(subjective claims and non-subjective claims).

\subsubsection{Subjective expression claims}
On social media, subjective expression claims often carry emotional tones and personal viewpoints, potentially including humor, satire, or cultural references. By considering these factors, we have designed corresponding prompt templates:
\begin{mybox}
\textbf{Instruction}: You need to do: 

(1) Evaluate whether the content of the source post is a reasonable subjective expression, considering context, humor, satire, and cultural references. 

(2) Evaluate whether the content of the original post may damage public trust in government or public figures.\\
\textbf{Claim}: [Claim]\\
\textbf{Comment}: [Comment]\\
At the end, please choose the answer from the following options: Fake, Real.
\end{mybox}
The first point, we designed Reasonability Prompts to enable agents to accurately determine whether a post reasonably expresses a personal opinion; the second point, we designed Trust Impact Prompts to help agents assess potential damage to public trust in government or public figures, identifying rumors that could undermine public confidence.

% \begin{algorithm}[tb]
% \caption{\themodel: Stance Separated Multi-Agent Debate}
% \label{alg}
% \begin{algorithmic}[1]
% \Require Debate claim $c$, comments $x_i^n$, number $k$, rounds $M$, debaters $N$
% \Ensure Final answer $\hat{y}$ 
% \Procedure{S2MAD}{$c, x_i^n, k, M, N$}
%     \State Initialize lists: $P$ (support), $N$ (oppose)
%     \For{$x_i$ in $x_i^n$} \Comment{Score comment}
%         \State $s_i \gets LLM(c, x_i)$
%         \State Add $(x_i, s_i)$ to $P$ if $s_i > 0$ else to $N$
%     \EndFor
%     \State $P \gets \text{top\_k}(P)$ \Comment{Select comments}
%     \State $N \gets \text{top\_k}(N)$
%     \State $A_P \gets LLM(c, P)$ \Comment{Initial Opinion}
%     \State $A_N \gets LLM(c, N)$
%     \State $J$ \Comment{Initialize judge}
%     \State $D \gets [D_1, D_2]$ \Comment{Initialize debaters}
%     \For{$m = 1$ to $M$}
%         \State $h_P \gets D_1(c, A_N)$, $h_N \gets D_2(c, A_P)$
%         \State $A_P \gets h_P$, $A_N \gets h_N$
%     \EndFor
%     \If{$h_P = h_N$}
%         \State $\hat{y} \gets h_P$ \Comment{Consensus reached}
%     \Else
%         \State $\hat{y} \gets J(A_P, A_N)$ \Comment{Judge decides}
%     \EndIf
%     \State \Return $\hat{y}$
% \EndProcedure
% \end{algorithmic}
% \end{algorithm}

\begin{algorithm}[tb]
\caption{\themodel: Stance Separated Multi-Agent Debate}
\label{alg}
\begin{algorithmic}[1]
\Require Debate claim $c$, Comments set $\mathcal{T}(c)$, Rounds $M$, Agents $D_p,D_n,J$
\Ensure Final prediction $\hat{y}$ 
\Procedure{S2MAD} {$c, \mathcal{T}(c), M$}
    
    \State // \textit{Stance Separation}
    \For{$x_i$ \textbf{in} $\mathcal{T}(c)$}
    \State $s_i\gets$  Score $x_i$ by LLM \hfill Eq. \ref{eq:score}
    \EndFor
    \State $P\gets$ Select $k$ comments with highest scores
    \State $N\gets$ Select $k$ comments with lowest scores
    \hfill Eq. \ref{eq:topk}
    \vspace{0.1cm}
    \State // \textit{Initial Opinion Generation}
    % \State $type(c) \gets$ Judgment c by $LLM$
    % \State Initialization strategy for type(c):
    % \State \hspace{\algorithmicindent} $h_p^0\gets$ Initialize agent $D_p$ with $P$
    % \State \hspace{\algorithmicindent} $h_n^0\gets$ Initialize agent $D_n$ with $N$
    \State $isSub \gets$ Get claim $c$'s subjectivity
    \State $h_p^0\gets$ Initialize $D_p$ with $P$ based on $isSub$
    \State $h_n^0\gets$ Initialize $D_n$ with $N$ based on $isSub$
    \hfill Eq. \ref{eq:init}
    \vspace{0.1cm}
    \State // \textit{Multi-Agent Debate}
    \For{$j = 1$ to $M$}
    \State $h_p^j \gets$ Update $D_p$'s opinion with $h_n^{j-1}$
    \State $h_n^j \gets$ Update $D_n$'s opinion with $h_p^{j-1}$
    \hfill Eq. \ref{eq:updat}
    \EndFor
    \If{$h_p = h_n$}
        \State \textbf{return} $h_p$
    \Else
        \State \textbf{return} Judge $J$'s determination \hfill Eq. \ref{eq:judge}
        % \State $\hat{y} \gets$Judge $J$ assess all arguments ($h_p^M, h_n^M$)\\ \hfill(Equation \ref{eq:judge})
    \EndIf
    % \State return $\hat y$
\EndProcedure
\end{algorithmic}
\end{algorithm}

\subsubsection{Non-subjective claims}
Non-subjective claims on social media are usually based on facts, data, or verifiable information. These claims require carefully designed prompt templates to ensure their authenticity and reliability:
\begin{mybox}
\textbf{Instruction}:You need to do:

(1) Evaluate the consistency and reliability of the supporting comments. Look for specific facts, data, or credible sources. 

(2) Assess the consistency and reliability of the rebuttal comments. Identify any valid points that raise doubts. 

(3) Consider common sense and general knowledge related to the topic.\\
\textbf{Claim}: [Claim]\\
\textbf{Comment}: [Comment]\\
At the end, please choose the answer from the following options: Fake, Real.
\end{mybox}
First, we created Support Evaluation Prompts to help agents focus on facts and credible sources. Second, we designed Rebuttal Evaluation Prompts to enable agents to critically analyze opposing viewpoints and uncover any potential flaws or discrepancies. Moreover, we introduced Common Sense Verification Prompts to encourage agents to use general knowledge and logical reasoning to assess the plausibility of claims, thereby effectively identifying misinformation.

We generate initial opinion for claims through agents, mathematically expressed as:
\begin{equation}\label{eq:init}
\begin{aligned}
    h_p^0=\text{LLM}(p_{io},c,P)\\
    h_n^0=\text{LLM}(p_{io},c,N)
\end{aligned}
\end{equation}
Where $p_{io}$ is the prompt used for generating the initial opinion, which refers to the guiding prompts for different types of claims mentioned above. $h_p^0$ represents the initial opinion based on the support stance comment set $P$, and $h_n^0$ denotes the initial opinion based on the opposing stance comment set $N$.

\subsection{Multi-Agent Debate}

In our method, instead of relying on a single agent to predict the veracity of claims, we use multiple agents to fully leverage their diverse knowledge. During the initial opinion generation phase, we prompt each agent to independently reason whether a given claim is a rumor by analyzing the statement text and comments with different stances. However, agent reasoning sometimes produces incorrect answers, and some recent studies ~\cite{gou2023critic, liang2023encouraging, zhang2024enhancing} have also shown that agents struggle to self-correct their responses without external feedback. To address this, we incorporate a debate scheme to integrate the responses from different agents. Generally, the framework is composed of two components, which are detailed as follows:

\subsubsection{Agent as Debater} Once we obtain the initial viewpoints, we designate the agents to participate in the debate. We have two agents referred to as debaters, denoted as $D=\{D_p, D_n\}$. In the first round of the debate, we prompt each debater to generate a response based on the initial opinion ($h_p^0$ and $h_n^0$) produced by the other debaters. In subsequent rounds, each debater continues to respond based on the debate history provided by the other debater until the maximum number of debate rounds $M$ is reached. Mathematically expressed as:
\begin{equation}
\begin{aligned}\label{eq:updat}
    h_p^j=\text{LLM}(p_{d},h_n^{j-1}), j \leq M\\
    h_n^j=\text{LLM}(p_{d}, h_p^{j-1}), j \leq M
\end{aligned}
\end{equation}
Where $h_i^j$ represents the response generated by debater$i$ in the j-th round of the debate, $p_{d}$ is the prompt used for the debate:

\begin{mybox}
\textbf{Instruction}: These are the opinions from other debaters. Based on the opinion of yourself and other debaters, you need: use critical thinking to analyze the views of others. \\
\textbf{Other debaters' opinions}: [Other answers]\\
Can you give an updated response, at the end, please choose the answer from the following options: Fake, Real.
\end{mybox}
In multi-round debates, to ensure that debaters are not stubborn and can deeply analyze and understand the logic and basis of arguments, we have included a critical prompt. This helps them better understand the topic of the debate and related issues, identify logical flaws, biases, or insufficient evidence in the opposing arguments, and thus provide stronger rebuttals.

\subsubsection{Agent as Judge} Furthermore, in multi-round debates where opinions do not converge to a consensus answer, it is crucial to designate an Judge, referred to as $J$, to deduce the final answer. Judge $J$ will assess all arguments (Agent$_p$'s last reply and Agent$_n$'s last reply) made throughout the debate, analyze their logical coherence and the level of evidence supporting them, and synthesize all the information to determine the most persuasive conclusion. Prompt $p_{j}$ designed for the judge  is as follows:

\begin{mybox}
\textbf{Instruction}: Please judge whether the post text is fake or real based on the following debate between $Agent_p$ and $Agent_n$:\\
\textbf{Claim}: [Claim]\\
\textbf{Agent$_p$ arguing}: [Agent$_p$'s last reply]\\
\textbf{Agent$_n$ arguing}: [Agent$_n$'s last reply]\\
Consider the arguments presented by both agents and make your determination about the authenticity of the post. At the end, must choose the answer from the following options: Fake, Real.
\end{mybox}

Mathematically expressed as:
\begin{equation}\label{eq:judge}
    \hat{y}=\text{LLM}(p_{j},h_p^{M},h_n^{M})
\end{equation}
where $\hat{y}$ denotes the consensus answer of debaters, and $h_p^{M}$ and $h_n^{M}$ represent the reasoning of the debaters in the final round of the debate. 

%% file: experiment.tex
\section{Experiment}
In this section, we evaluate the proposed methods using widely used~\cite{lin2022detect,lin2023zero,zhang2024t3rd} real-world social media rumor datasets related to the breaking event of COVID-19, namely Twitter-COVID19 and Weibo-COVID19. We conducted extensive experiments to validate the effectiveness of our approach.

\begin{table*}[htbp]
  \centering
    \renewcommand\arraystretch{1.2}
    \tabcolsep=0.1cm
    \begin{tabular}{p{3cm}|cccc|cccc}
    \toprule
    \multirow{2}[2]{*}{\textbf{Models}} & \multicolumn{4}{c|}{\textbf{Twitter-COVID19}} & \multicolumn{4}{c}{\textbf{Weibo-COVID19}} \\
          & ACC.  & Mac-F1 & RF1   & NF1   & ACC.  & Mac-F1 & RF1   & NF1 \\
    \midrule
    BiGCN & 0.593 & 0.557 & 0.647 & 0.468 & 0.619 & 0.534 & 0.720 & 0.349 \\
    ACLR-BiGCN & 0.692 & 0.614 & 0.785 & 0.444 & 0.670 & 0.594 & 0.754 & 0.435 \\
    RPL   & 0.727 & 0.697 & 0.793 & 0.601 & 0.745 & 0.719 & 0.804 & 0.634 \\
    T$^3$RD  & 0.735 & 0.701 & 0.808 & 0.593 & 0.797 & \textbf{0.788} &0.832 & 0.743 \\
    \midrule
    Qwen1.5 &   0.565    &   0.565    &   0.580    &   0.549    & 0.698 & 0.670 &0.762  &0.578  \\
    Qwen1.5 + S2MAD\  & 0.645\textit{\small +.080} & 0.645\textit{\small +.080} & 0.634\textit{\small +.054} & \textbf{0.655}\textit{\small +.106} & 0.737\textit{\small +.039} & 0.757\textit{\small +.087} & 0.765\textit{\small +.003} & \textbf{0.748}\textit{\small +.170} \\
    \midrule
    GPT3.5 & 0.643 & 0.567 & 0.749 & 0.384 & 0.762 & 0.682 &  0.843 & 0.520 \\
    GPT3.5 + S2MAD  & \textbf{0.765}\textit{\small +.122} & \textbf{0.737}\textit{\small +.170} & \textbf{0.827}\textit{\small +.078} & 0.647\textit{\small +.263} & \textbf{0.810}\textit{\small +.048} & 0.764\textit{\small +0.082} & \textbf{0.869}\textit{\small +.026} & 0.658\textit{\small +.138} \\
    \bottomrule
    \end{tabular}%
\caption{
  Rumor detection results on the target datasets. GPT3.5 refers to the GPT-3.5 Turbo model, and Qwen1.5 refers to the Qwen1.5-7b-chat model.\\
  }
  \label{tab:main results}
\end{table*}%

\subsection{Datasets}
We use breaking events related datasets, Twitter-COVID19~\cite{lin2022detect} and Weibo-COVID19~\cite{lin2022detect}, to evaluate the \themodel approach. These datasets are sourced from two popular social media platforms—Twitter and Weibo. The Twitter-COVID19 dataset consists of English rumor datasets with conversation threads in tweets, providing a rich context for analysis. On the other hand, the Weibo-COVID19 dataset comprises Chinese rumor datasets with a similar composition structure. These datasets are annotated with two labels: Rumor and Non-Rumor, which are used for the binary classification of rumors and non-rumors.

We employ a range of commonly used evaluation metrics to validate the effectiveness of the proposed method. Firstly, Accuracy (ACC.) measures the proportion of correctly predicted samples, reflecting overall prediction precision. Secondly, the F1-score integrates Precision and Recall, comprising Positive F1 (RF1), Negative F1 (NF1), and Macro-average F1 (Mac-F1), assessing performance for different classes and overall average performance.

\subsection{Settings}
We compared our model with several baseline models: the GCN-based model BiGCN~\cite{bian2020rumor}, the adversarial contrastive learning framework ACLR-BiGCN~\cite{lin2022detect} built on top of BiGCN, the zero-shot rumor detection model based on prompt learning RPL~\cite{lin2023zero}, and the test-time training model for rumor detection T$^3$RD~\cite{zhang2024t3rd}. In our experiments, to adapt to real-world breaking events, we tested the BiGCN, ACLR-BiGCN, and T$^3$RD models in a zero-shot setting. These models were trained on public datasets(Twitter dataset~\cite{ma2017detect} or Weibo dataset~\cite{ma2016detecting}) and then directly tested on target datasets to evaluate their performance and adaptability. This approach allows for a better assessment of the models' effectiveness in handling breaking events. Additionally, we used Qwen1.5-7b-chat and GPT-3.5 turbo as our baseline models under zero-shot settings. Specifically, for each sample, we randomly selected 40 comments and used Vanilla Prompt~\cite{liu2024can} for inference.

 \themodel was performed in a zero-shot setting, requiring no training data, using all samples from the evaluation dataset. During the stance separation phase, we utilized glm-3-turbo as our scorer. Initially, we employed Qwen1.5-7b-chat as the base debater model for our experiments. Additionally, we validated \themodel's performance by conducting experiments with GPT-3.5 turbo as the debater.
In our experiments, the open-source Qwen1.5-7b-chat model was run on a remote machine server with 2 NVIDIA RTX 3090 (24G) GPUs. During the experiments, to accommodate the contentious nature of opinions in breaking events, we set the temperature to 0.2, while all other hyperparameters for generating LLM outputs were kept at their default settings.

\subsection{Overall Performance}
Table \ref{tab:main results} shows the comparison of our method with various baseline models on two real-world datasets related to breaking events, Twitter-COVID19 and Weibo-COVID19. The data clearly indicates that \themodel consistently and significantly improves model performance across both datasets, regardless of whether Qwen1.5-7b-chat or GPT-3.5 turbo is used as the backbone model.

Leveraging the strong reasoning capabilities of the propagation graph structure, the four graph-based detection models exhibit good performance, with T3RD showing the best results. Our results with the two backbone models, Qwen1.5-7b-chat and GPT-3.5 turbo, reveal that LLMs, when applied directly to social media rumor detection during breaking events in a zero-shot setting, do not demonstrate strong reasoning capabilities. This finding corroborates our earlier discussion. However, when our method is applied to these two backbone models, their performance improves significantly. Specifically, on the Twitter-COVID19 dataset, \themodel enhances Qwen1.5-7b-chat by 8\% and GPT-3.5 turbo by 12.2\%. On the Weibo-COVID19 dataset, \themodel boosts Qwen1.5-7b-chat by 3.9\% and GPT-3.5 turbo by 4.8\%. Furthermore, our proposed \themodel method achieves the best performance when used with the GPT-3.5 turbo backbone model. These results strongly demonstrate the superiority of \themodel.

\subsection{Ablation Study}
\subsubsection{Debate Rounds}

\begin{figure}
    \centering
    \includegraphics[width=0.9\linewidth]{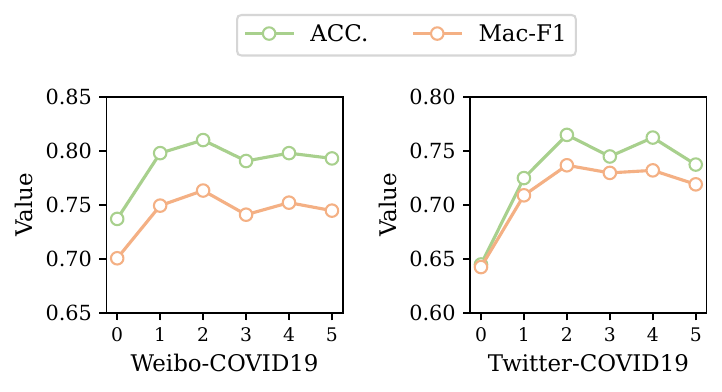}
    \vspace{-0.3cm}
    \caption{Comparison of different numbers of debate rounds in our proposed \themodel approach using GPT-3.5 turbo.}
    \label{fig:Debate Rounds}
    
\end{figure}

First, we analyzed the impact of debate rounds in \themodel using GPT-3.5 turbo. In Figure \ref{fig:Debate Rounds}, we increased the number of debate rounds while keeping the number of debaters fixed at two. We found that when the number of debate rounds did not exceed two, \themodel showed a significant performance improvement on the Twitter-COVID19 and Weibo-COVID19 datasets. However, we observed that when the number of debate rounds exceeded two, the accuracy of the model on both datasets showed slight fluctuations, and the final performance was similar to that of two rounds of debate. We speculate that after the second round, the debaters have largely reached a consensus on the veracity of most samples, and only on a few samples where consensus was not reached did the Judge show unstable responses based on the debaters' fully expressed viewpoints. Therefore, we default to conducting two rounds of debate.

\subsubsection{Effectiveness of Stance Separation}

We examined the impact of stance separation in \themodel using GPT-3.5 turbo. To ensure the validity of the validation, we no longer filtered comments based on their support or opposition to the claims. Instead, we randomly sampled up to forty comments from each claim, divided them into two equal parts, and assigned them to two agents, thereby removing any specific stance bias. The agents then generated initial perspectives based on their respective sets of comments, while other experimental settings remained consistent with \themodel. As shown in Table \ref{tab:ablation}, removing stance separation led to a decline in \themodel's performance across both datasets. These results clearly demonstrate the effectiveness of stance separation in improving the performance of \themodel, thereby supporting our assertion that categorizing comments based on their stance towards the original claim allows the model to better understand and process the information, leading to more accurate rumor detection.

\begin{table}[tbp]
  \centering
    \begin{tabular}{c|c|c}
    \toprule
       \textbf{Model}   & \textbf{Twitter-COVID19} & \textbf{Weibo-COVID19} \\
    \midrule
     \textbf{Full Model} & \textbf{0.765} & \textbf{0.810} \\
    \midrule
    w/o Stance & 0.648 & 0.674 \\
    % \midrule
    w/o Non-Sub  & 0.575 & 0.801 \\
    w/o Sub & 0.703 & 0.706 \\
    % \midrule
    w/o Debate & 0.645 & 0.737 \\
    \bottomrule
    \end{tabular}%
  \caption{Experimental results of \themodel with different settings using GPT-3.5 turbo. ``w/o Stance" indicates no stance separation. ``w/o Non-Sub/Sub'' indicates applying the initial opinion prompt corresponding to subjective/non-subjective claims to the entire dataset. ``w/o Debate'' indecates no debate process.}
  \label{tab:ablation}%
\end{table}%

\subsubsection{Effectiveness of Categorizing the Claim}
Further, we extensively studied the impact of categorizing claims into subjective and non-subjective types in our approach using GPT-3.5 turbo. Specifically, during the initial opinion generation phase, we stopped distinguishing between subjective and non-subjective claims and instead used the initial opinion prompts for both types across the entire dataset.
Our experimental results, presented in Table \ref{tab:ablation}, show that the full model achieved an accuracy of 0.765 on the Twitter-COVID19 dataset and 0.810 on the Weibo-COVID19 dataset. When the initial opinion prompt for non-subjective claims was removed (``w/o Non-Sub"), the accuracy dropped to 0.575 on Twitter-COVID19 and 0.801 on Weibo-COVID19. Similarly, removing the prompt for subjective claims (``w/o Sub") resulted in accuracies of 0.703 on Twitter-COVID19 and 0.706 on Weibo-COVID19.
These findings demonstrate that categorizing claims into subjective and non-subjective types significantly enhances our model's performance. By using appropriate prompts for each claim type, our approach achieves better results in social media rumor detection during breaking events.

\subsubsection{Effectiveness of Debate}

We investigated the effectiveness of incorporating debate in our rumor detection approach. Specifically, keeping other experimental settings the same, we compared the performance of our complete approach architecture with a single-agent reasoning model to evaluate the impact of debate on our approach. In Table \ref{tab:ablation}, ``/o Debate" denotes the removal of the debate component, where a single agent performs reasoning under the same conditions. As shown, removing the debate from \themodel led to a decrease in accuracy of 12\% on the Twitter-COVID19 dataset and 7.3\% on the Weibo-COVID19 dataset.
These findings clearly demonstrate the effectiveness of the multi-agent debate in improving the performance of our model in social media rumor detection during breaking events. This supports our earlier assertion that the debate process allows for a more thorough examination of claims by considering multiple perspectives, leading to more accurate and reliable detection outcomes.

\subsection{Early Detection}
\begin{figure}
    \centering
    \includegraphics[width=0.98\linewidth]{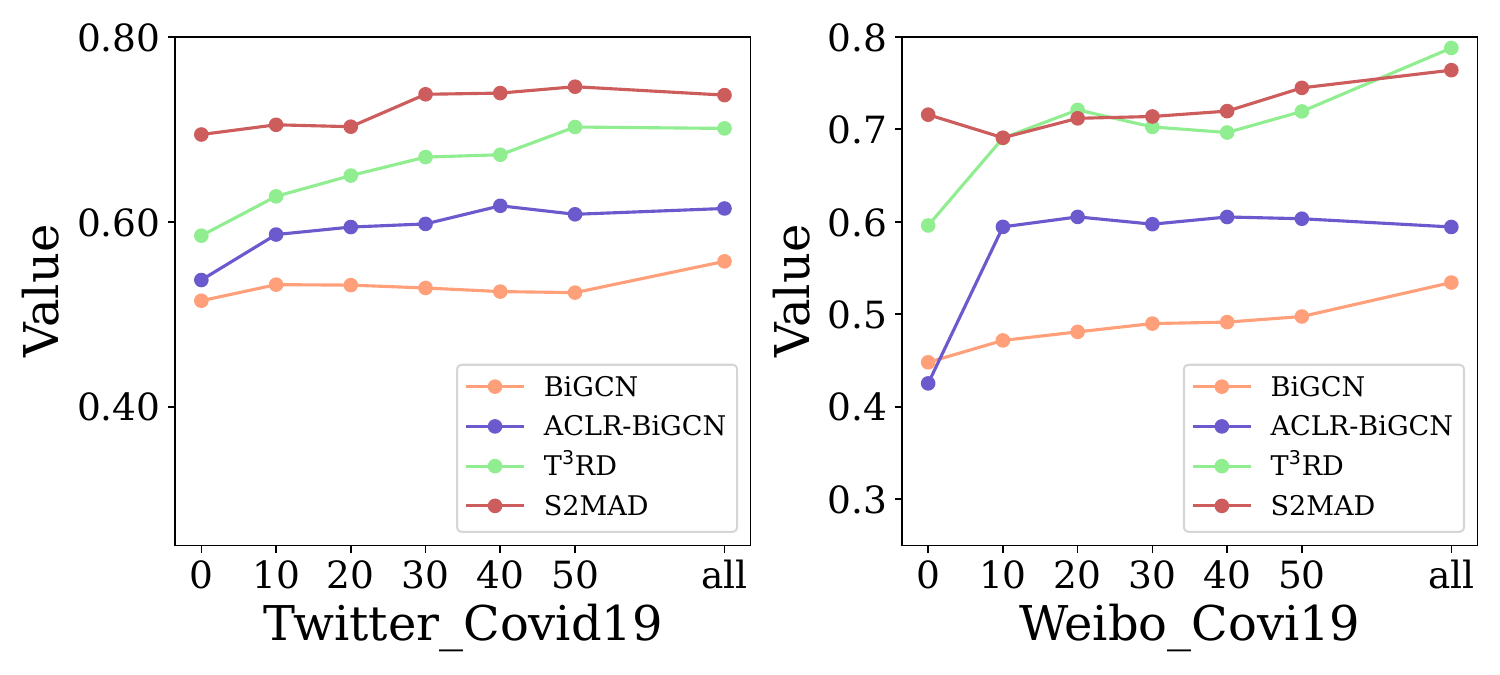}
    \vspace{-10pt}
       \caption{Early detection performance is evaluated at different checkpoints based on the post count.}
       \label{fig:early}
\end{figure}

Early detection plays a crucial role in effectively preventing the widespread dissemination of rumors, making it an important metric for model evaluation. By setting different checkpoints, models are assessed using the post count up to each detection point. The performance of each model is evaluated using the Macro F1 score. Figure \ref{fig:early} illustrates the early detection performance of our model compared to others at various checkpoints. Notably, our model maintains competitive or superior performance across different checkpoints on both datasets. This indicates that our approach is particularly adept at identifying potential rumors before they gain significant traction, thereby providing a valuable tool in combating the spread of false information online.

%% file: conclusion.tex
\section{Conclusion}
In this paper, we investigate the use of multi-agent debate for rumor detection in social media within the context of breaking event scenarios. We propose an approach called \themodel. \themodel uses LLMs to separate comments based on their stance towards the claim. Furthermore, we categorize claims into subjective and non-subjective types and design different guide prompts to direct agents in reasoning about the veracity of the claims based on comments with different stances to generate initial opinions. Subsequently, we conduct a multi-agent debate based on these initial opinions. Finally, after reaching the maximum number of debate rounds, if the agents have not reached a consensus on the veracity of the claims, we employ a designed Judge to provide the final conclusion. We validated the effectiveness of \themodel through extensive experiments on two datasets related to breaking event scenarios, Weibo-COVID19 and Twitter-COVID19.